\documentclass[11pt,a4paper]{article}
\usepackage[nohyperref]{naaclhlt2019}
\usepackage{times}
\usepackage{latexsym}
\usepackage{multirow}
\usepackage{url}
\usepackage{graphicx}
\usepackage{bm}
\usepackage{amsmath}
\usepackage{amssymb}
\usepackage{array}

\aclfinalcopy 
\def\aclpaperid{423} 

\title{Distant Supervision Relation Extraction with Intra-Bag \\
and Inter-Bag Attentions}

\author{
  Zhi-Xiu Ye \\
  University of Science and \\
  Technology of China \\
  {\tt zxye@mail.ustc.edu.cn} \\\And
  Zhen-Hua Ling \\
  University of Science and \\
  Technology of China \\
  {\tt zhling@ustc.edu.cn} \\
  }

\date{}

\begin{document}
\maketitle
\begin{abstract}
This paper presents a neural relation extraction method to deal with the noisy training data generated by distant supervision.
Previous studies mainly focus on sentence-level de-noising by designing neural networks with intra-bag attentions.
In this paper, both intra-bag and inter-bag attentions are considered  in order to deal with the noise at sentence-level and bag-level respectively.
First, relation-aware bag representations are calculated by weighting sentence embeddings using intra-bag attentions.
Here, each possible relation is utilized as the query for attention calculation instead of only using the target relation in conventional methods.
Furthermore, the representation of a group of bags in the training set  which share the same relation label is calculated by
weighting bag representations using a similarity-based inter-bag attention module.
Finally, a bag group is utilized as a training sample when building our relation extractor.
Experimental results on the New York Times dataset demonstrate the effectiveness of our proposed intra-bag and inter-bag attention modules.
Our method also achieves better relation extraction accuracy than state-of-the-art  methods on this dataset\footnote{The code is available at \url{https://github.com/ZhixiuYe/Intra-Bag-and-Inter-Bag-Attentions}.}.
\end{abstract}

\section{Introduction}
\label{sec:intro}
\noindent Relation Extraction is a fundamental task in natural language processing (NLP), which aims to extract semantic relations between entities.
For example, sentence ``[\emph{Barack Obama}]$_{e1}$ was born in [\emph{Hawaii}]$_{e2}$'' expresses the relation \emph{BornIn} between entity pair \textbf{Barack Obama} and \textbf{Hawaii}.

\begin{table}
\small
\centering
\begin{tabular}{p{0.4cm}|p{5.0cm}|p{0.9cm}}
\hline
\bf bag &\bf sentence &\bf correct? \\ \hline
{\multirow{4}{*}{B1}}&S1. \textbf{Barack Obama} was born in \textbf{the United States}.& {\multirow{2}{*}{\bf Yes}} \\ \cline{2-3}
&S2. \textbf{Barack Obama} was the 44th president of \textbf{the United States} & {\multirow{2}{*}{\bf No}}  \\ \hline
{\multirow{7}{*}{B2}}&S3. \textbf{Kyle Busch} , a \textbf{Las Vegas} resident who ran second to Johnson last year, finished third, followed by Kasey Kahne, Jeff Gordon and mark martin . & {\multirow{4}{*}{\bf No}} \\ \cline{2-3}
&S4. Hendrick drivers finished in three of the top four spots at \textbf{Las Vegas} , including \textbf{Kyle Busch} in second and ... & {\multirow{3}{*}{\bf No}}  \\ \hline
\end{tabular}
\caption{\label{examples} Examples of sentences with relation \emph{place\_of\_birth} annotated by distant supervision, where ``Yes" and ``No" indicate whether or not
each sentence actually expresses this relation.}
\end{table}

Conventional relation extraction methods, such as \citep{zelenko2003kernel,culotta2004dependency,mooney2006subsequence}, adopted supervised training and suffered from the lack of large-scale manually labeled data.
To address this issue, the distant supervision method \citep{mintz2009distant} was proposed, which generated the data for training relation extraction models automatically.
The distant supervision assumption says that if two entities participate in a relation, \textbf{all} sentences that mention these two entities express that relation.
It is inevitable that there exists noise in the data labeled by distant supervision.
For example, the precision of aligning the relations in Freebase to the New York Times corpus was only about 70\% \citep{riedel2010modeling}.

Thus, the relation extraction method proposed in  \citep{riedel2010modeling} argued that the distant supervision assumption was too strong and relaxed it to \emph{expressed-at-least-once} assumption.
This assumption says that if two entities participate in a relation, \textbf{at least one sentence} that mentions these two entities might express that relation.
An example is shown by sentences S1 and S2 in Table \ref{examples}.
This relation extraction method first divided the training data given by distant supervision into bags
where each bag was a set of sentences containing the same entity pair.
Then, bag representations were derived by weighting sentences within each bag.
It was expected that the weights of the sentences with incorrect labels were reduced and the bag representations were calculated mainly using the sentences with correct labels.
Finally, bags were utilized as the samples for training relation extraction models instead of sentences.

In recent years, many relation extraction methods using neural networks with attention mechanism \citep{lin2016neural,ji2017distant,jat2018improving} have been proposed
to alleviate the influence of noisy training data under the \emph{expressed-at-least-once}  assumption.
However, these methods still have two deficiencies.
First, only the target relation of each bag is used to calculate the attention weights for
deriving bag representations from sentence embeddings at training stage.
Here we argue that the bag representations should be calculated in a relation-aware way.
For example, the bag B1 in Table \ref{examples} contains two sentences S1 and S2.
When this bag is classified to relation \emph{BornIn}, the sentence S1 should have higher weight than S2,
but when classified to relation \emph{PresidentOf}, the weight of S2 should be higher.
Second, the \emph{expressed-at-least-once} assumption ignores the \textbf{noisy bag problem} which means that all sentences in one bag are incorrectly labeled.
An example is shown by bag B2 in Table \ref{examples}.

In order to deal with these two deficiencies of previous methods, this paper proposes a neural network with multi-level attentions for distant supervision relation extraction.
At the instance/sentence-level, i.e., intra-bag level, all possible relations are employed as queries to calculate the relation-aware bag representations
instead of only using the target relation of each bag.
To address the noisy bag problem, a bag group is adopted as a training sample instead of a single bag.
Here, a bag group is composed of bags in the training set which share the same relation label.
The representation of a bag group is calculated by weighting bag representations using a similarity-based inter-bag attention module.

The contributions of this paper are threefold.
First, an improved intra-bag attention mechanism is proposed to derive relation-aware bag representations for relation extraction.
Second, an inter-bag attention module is introduced to deal with the noisy bag problem which is ignored by the expressed-at-least-once assumption.
Third, our  methods achieve better extraction accuracy than state-of-the-art models on the widely used New York Times (NYT) dataset \citep{riedel2010modeling}.

\section{Related Work}
Some previous work \citep{zelenko2003kernel,mooney2006subsequence} treated relation extraction as a supervised learning task and designed hand-crafted features to train kernel-based models.
Due to the lack of large-scale manually labeled data for supervised training, the distant supervision approach \citep{mintz2009distant} was proposed,
which aligned raw texts toward knowledge bases automatically to generate relation labels for entity pairs.
However, this approach suffered from the issue of noisy labels.
Therefore, some subsequent studies \citep{riedel2010modeling,hoffmann2011knowledge,surdeanu2012multi} considered distant supervision relation extraction as a multi-instance learning problem, which extracted relation from a bag of sentences instead of a single sentence.

\begin{figure*}[!t]
\centering
\includegraphics[width=6.2in]{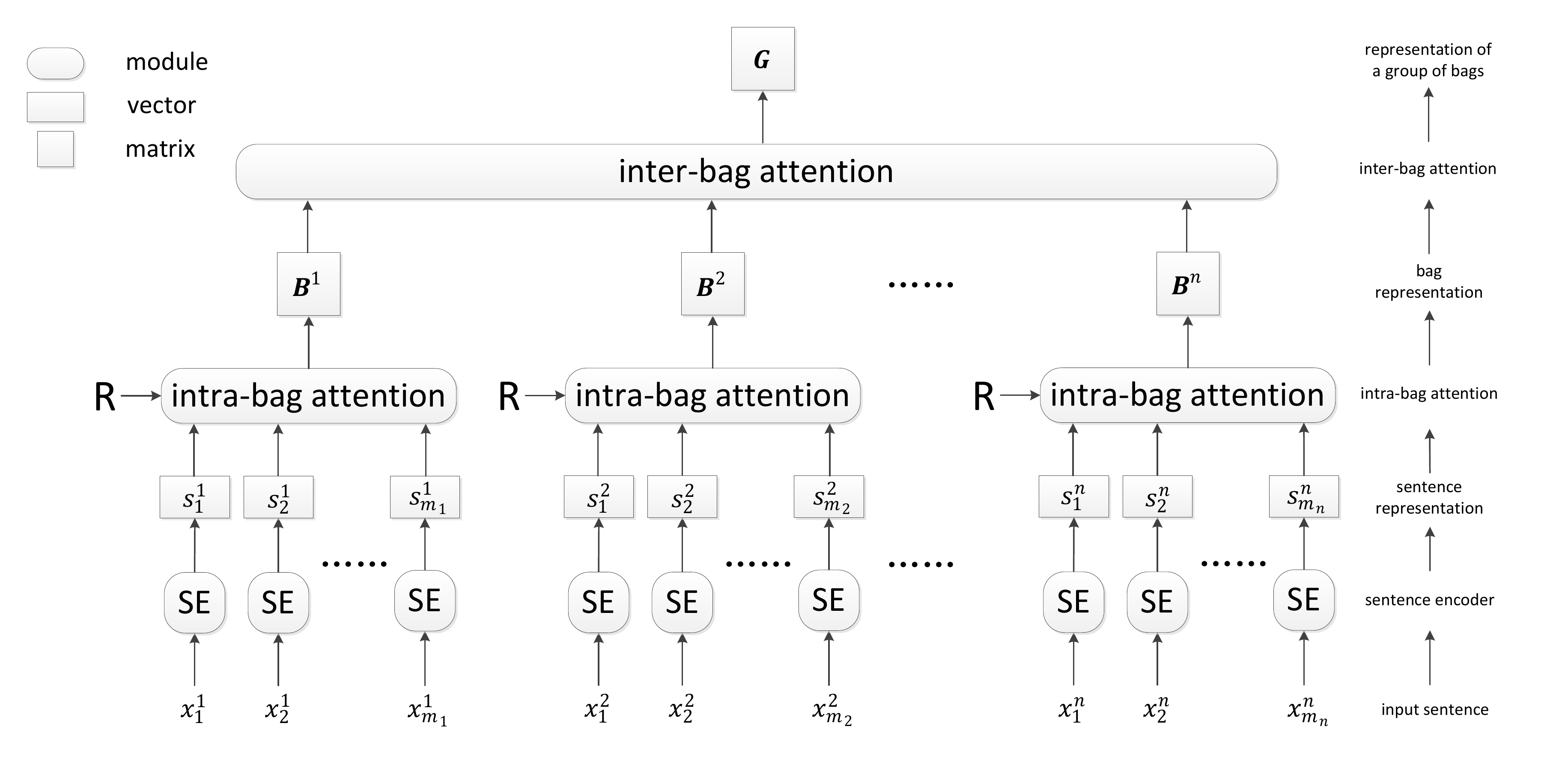}
\caption{
The framework  of our proposed neural network with intra-bag and inter-bag attentions for relation extraction.
}
\label{framework}
\end{figure*}

With the development of deep learning techniques \citep{lecun2015deep}, many neural-network-based models have been developed for distant supervision relation extraction.
\citet{zeng2015distant} proposed piecewise convolutional neural networks (PCNNs) to model sentence representations and chose the most reliable sentence as the bag representation.
\citet{lin2016neural} employed PCNNs as sentence encoders and proposed an intra-bag attention mechanism to compute the bag representation via a weighted sum of all sentence representations in the bag.
\citet{ji2017distant} adopted a similar attention strategy and combined entity descriptions to calculate the weights.
\citet{liu2017soft} proposed a soft-label method to reduce the influence of noisy instances.
All these methods represented a bag with a weighted sum of sentence embeddings, and calculated the probability of the bag being classified into each relation using the same bag representation at training stage.
In our proposed method, intra-bag attentions are computed in a relation-aware way,
which means that different bag representations are utilized to calculate the probabilities for different relation types.
Besides, these existing methods focused on intra-bag attentions and ignored the noisy bag problem.

Some data filtering strategies for robust distant supervision relation extraction have also been proposed.
\citet{feng2018reinforcement} and \citet{qin2018robust} both employed reinforcement learning to train instance selector and to filter out the samples with wrong labels.
Their rewards were calculated from the prediction probabilities and the performance change of the relation classifier respectively.
\citet{qin2018dsgan} designed an adversarial learning process to build a sentence-level generator via policy-gradient-based reinforcement learning.
These methods were proposed to filter out the noisy data at sentence-level and also failed to deal with the noisy bag problem explicitly.

\section{Methodology}
\noindent In this section, we introduce a neural network with intra-bag and inter-bag attentions for distant supervision relation extraction.
Let $g = \{ b^{1}, b^{2}, ..., b^{n} \}$ denote a group of bags which have the same relation label given by distant supervision, and $n$ is the number of bags within this group.
Let $b^{i} = \{ x^{i}_{1}, x^{i}_{2}, ..., x^{i}_{m_{i}} \}$ denote all sentences in bag $b^{i}$, and $m_{i}$ is the  number of sentences in bag $b^{i}$.
Let $x^{i}_{j} = \{ w^{i}_{j1}, w^{i}_{j2}, ..., w^{i}_{jl_{ij}} \}$ denote the $j$-th sentence in the $i$-th bag and $l_{ij}$ is its length (i.e., number of words).
The framework of our model is shown in Fig. \ref{framework}, which has three main modules.
\begin{itemize}
\item \textbf{Sentence Encoder} Given a sentence $x^{i}_{j}$ and the positions of two entities within this sentence,
CNNs or PCNNs \citep{zeng2015distant} are adopted to derive the sentence representation $\mathbf{s}^{i}_{j}$.
\item \textbf{Intra-Bag Attention} Given the sentence representations of all sentences within a bag $b^{i}$ and a relation embedding matrix $\mathbf{R}$,
attention weight vectors $\boldsymbol{\alpha}^{i}_{k}$ and bag representations $\mathbf{b}^{i}_{k}$ are calculated for all relations, where $k$ is the relation index.
\item \textbf{Inter-Bag Attention} Given the representations of all bags with the group $g$, a weight matrix $\boldsymbol{\beta}$ is further calculated via similarity-based attention mechanism to obtain the representation of the bag group.
\end{itemize}
More details of these three modules will be introduced in the following subsections.

\subsection{Sentence Encoder}
\subsubsection{Word Representation}
\noindent Each word $w^{i}_{jk}$ within the sentence $x^{i}_{j}$  is first mapped into a $d_{w}$-dimensional
word embedding.
To describe the position information of two entities, the position features (PFs) proposed in \citep{zeng2014relation} are also adopted in our work.
For each word, the PFs describe the relative distances between current word and the two entities and are further mapped into two vectors $\mathbf{p}^{i}_{jk}$ and $\mathbf{q}^{i}_{jk}$ of  $d_{p}$ dimensions.
Finally, these three vectors are concatenated to get the word representation  $\mathbf{w}^{i}_{jk} = [\mathbf{e}^{i}_{jk}; \mathbf{p}^{i}_{jk}; \mathbf{q}^{i}_{jk}]$  of  $d_{w} + 2d_{p}$ dimensions.

\subsubsection{Piecewise CNN}
\noindent
For sentence $x^{i}_{j}$, the matrix of word representations $\mathbf{W}^{i}_{j} \in \mathbb{R}^{l_{ij} \times (d_{w} + 2d_{p})}$ is first input into a CNN with $d_{c}$ filters.
Then, piecewise max pooling \citep{zeng2015distant} is employed to extract features from the three segments of CNN outputs, and
the segment boundaries are determined by the positions of the two entities. Finally, the sentence representation $\mathbf{s}^{i}_{j} \in \mathbb{R}^{3d_{c}}$ can be obtained.

\subsection{Intra-Bag Attention}
\noindent Let  $\mathbf{S}^{i} \in \mathbb{R}^{m_{i} \times 3d_{c}}$ represent the representations of all sentences within bag $b^{i}$,
and $\mathbf{R} \in \mathbb{R}^{h \times 3d_{c}}$ denote a relation embedding matrix where $h$ is the number of relations.

Different from conventional methods \citep{lin2016neural,ji2017distant} where a unified bag representation was derived for relation classification,
our method calculates bag representations $\mathbf{b}^{i}_{k}$ for bag $b^{i}$ on the condition of all possible relations as
\begin{equation}
\mathbf{b}^{i}_{k} = \sum^{m_{i}}_{j=1} \alpha^{i}_{kj} \mathbf{s}^{i}_{j},
\end{equation}
where $k\in\{1,2,...,h\}$ is the relation index and $\alpha^{i}_{kj}$ is the attention weight
between the $k$-th relation and the $j$-th sentence  in bag $b^{i}$.
$\alpha^{i}_{kj}$ can be further defined as
\begin{equation}
\alpha^{i}_{kj} = \frac{exp(e^{i}_{kj})}{\sum_{j^{'}=1}^{m_{i}} exp(e^{i}_{kj^{'}})},
\end{equation}
where $e^{i}_{kj}$ is the matching degree between the $k$-th relation query and the $j$-th sentence  in bag $b^{i}$.
In our implementation, a simple dot product between vectors is adopted to calculate the matching degree as
\begin{equation}
e^{i}_{kj} = \mathbf{r}_{k}
\mathbf{s}^{i\top}_{j},
\label{sentweight}
\end{equation}
where $\mathbf{r}_{k}$ is the $k$-th row of the relation embedding matrix $\mathbf{R}$\footnote{We also tried $\mathbf{r}_{k} \mathbf{A}
\mathbf{s}^{i\top}_{j}$, where $\mathbf{A}$ was a diagonal matrix,  in experiments and achieved similar performance.}.

Finally, the representations of bag $b^{i}$ compose the matrix $\mathbf{B}^{i} \in \mathbb{R}^{h \times 3d_{c}}$ in Fig. \ref{framework},
 where each row corresponds to a possible relation type of this bag.

\subsection{Inter-Bag Attention}
In order to deal with the noisy bag problem,
a similarity-based inter-bag attention module is designed to reduce the weights of noisy bags dynamically.
Intuitively, if two bags $b^{i_{1}}$ and $b^{i_{2}}$ are both labeled as relation $k$,
their representations $\mathbf{b}^{i_{1}}_{k}$ and $\mathbf{b}^{i_{2}}_{k}$ should be close to each other.
Given a group of bags with the same relation label, we assign higher weights to those bags which are close to other bags in this group.
As a result, the representation of bag group $g$ can be formulated as
\begin{equation}
\label{weightedbag}
\mathbf{g}_{k} = \sum^{n}_{i=1} \beta_{ik} \mathbf{b}^{i}_{k},
\end{equation}
where $\mathbf{g}_{k}$ is the $k$-th row of the matrix $\mathbf{G} \in \mathbb{R}^{h \times 3d_{c}}$ in Fig. \ref{framework}, $k$ is the relation index and $\beta_{ik}$ composes the attention weight matrix $\boldsymbol{\beta} \in \mathbb{R}^{n \times h}$.
Each $\beta_{ik}$ is defined as
\begin{equation}
\beta_{ik} = \frac{exp(\gamma_{ik})}{\sum_{i^{'}=1}^{n} exp(\gamma_{i^{'}k})},
\end{equation}
where $\gamma_{ik}$ describes the confidence of labeling bag $b^i$ with the $k$-th relation.

Inspired by the self-attention algorithm \citep{vaswani2017attention} which calculates the attention weights for a group of vectors using the vectors themselves,
we calculate the weights of bags according to their own representations.
Mathematically, $\gamma_{ik}$ is defined as
\begin{equation}
\gamma_{ik} = \sum_{i^{'}=1, ..., n, i^{'} \neq i} \mathrm{similarity}(\mathbf{b}^{i}_{k}, \mathbf{b}^{i^{'}}_{k}),
\end{equation}
where the function $\emph{similarity}$ is a simple dot product in our implementation as
\begin{equation}
\mathrm{similarity}(\mathbf{b}^{i}_{k}, \mathbf{b}^{i^{'}}_{k}) = {\mathbf{b}^{i}_{k}}\mathbf{b}^{i^{'}\top}_{k}.
\end{equation}
And also, in order to prevent the influence of vector length, all bag representations $\mathbf{b}^{i}_{k}$
are normalized to unit length as
$\overline{\mathbf{b}^{i}_{k}} = \mathbf{b}^{i}_{k} / ||\mathbf{b}^{i}_{k}||_{2}$
before calculating Eq.(4)-(7).

Then, the score $o_{k}$ of classifying bag group $g$ into relation $k$ is calculated via $\mathbf{g}_{k}$ and relation embedding $\mathbf{r}_{k}$ as
\begin{equation}
o_{k} = \mathbf{r}_{k}\mathbf{g}_{k}^{\top} + d_{k},
\label{dot}
\end{equation}
where $d_{k}$ is a bias term.
Finally, a softmax function is employed to obtain the probability that the bag group $g$ is classified into the $k$-th relation as
\begin{equation}
\label{probability}
{p}(k|g) = \frac{exp(o_{k})}{\sum_{k^{'}=1}^{h} exp(o_{k^{'}})}.
\end{equation}
It should be noticed that the same relation embedding matrix $\mathbf{R}$ is used for calculating Eq.(3) and Eq.(8).
Similar to \citet{lin2016neural}, the dropout strategy \citep{srivastava2014dropout} is applied to bag representation $\mathbf{B}^{i}$ to prevent overfitting.

\subsection{Implementation Details}
\subsubsection{Data Packing}
First of all, all sentences in the training set that contain the same two entities are accumulated into one bag.
Then, we tie up every $n$ bags that share the same relation label into a group.
It should be noticed that a bag group  is one training sample in our method.
Therefore, the model can also be trained in mini-batch mode by packing multiple bag groups into one batch.

\subsubsection{Objective Function and Optimization}
In our implementation, the objective function is defined as
\begin{equation}
\mathrm{J}(\theta) = - \sum_{(g,k) \in T} \mathrm{logp}(k|g;\theta)
\end{equation}
where $T$ is the set of all training samples and $\theta$ is the set of model parameters, including word embedding matrix, position feature embedding matrix, CNN weight matrix and relation embedding matrix.
The model parameters are estimated by minimizing the objective function J($\theta$) through mini-batch stochastic gradient descent (SGD).

\subsubsection{Training and Test}
As introduced above, at the training phase of our proposed method,  $n$ bags which have the same relation label are accumulated into one bag group and  the weighted sum of bag representations is calculated to obtain the  representation $\mathbf{G}$ of the bag group.
Due to the fact that the label of each bag is unknown at test stage, each single bag is treated as a bag group (i.e., $n$=1) when processing the test set.

And also, similar to \citep{qin2018robust}, we only apply inter-bag attentions to positive samples, i.e., the bags whose relation label is not $NA$ ($NoRelation$).
The reason is that the representations of the bags that express no relations are always  diverse and it's difficult to calculate suitable weights for them.

\subsubsection{Pre-training Strategy}
In our implementation, a pre-training strategy is adopted. We first train the model  with only intra-bag attentions until convergence.
Then, the inter-bag attention module is added and the model parameters are further updated until convergence again.
Preliminary experimental results showed that this strategy can lead to better model performance than considering inter-bag attentions from the very beginning.

\begin{table}[t!]
\small
\begin{center}
\begin{tabular}{|p{1.0in}<{\centering}|p{1.1in}<{\centering}|p{0.3in}<{\centering}|}
\hline
\bf Component                                       & \bf Parameter & \bf Value\\ \hline
word embedding & dimension & 50\\ \hline
{\multirow{2}{*}{position feature}}                 & max relative distance   & {\multirow{1}{*}{$\pm$30}}        \\ \cline{2-3}
                                                    & dimension & 5       \\ \hline
{\multirow{2}{*}{CNN}}                 & window size   & 3        \\ \cline{2-3}
                                                    & filter number & 230       \\ \hline
dropout                        & dropout rate  & 0.5      \\ \hline
{\multirow{6}{*}{optimization}}                                                     & strategy      & SGD     \\ \cline{2-3}
& learning rate & 0.1     \\ \cline{2-3}
                                                    & batch size $N_{p}$   & 50       \\ \cline{2-3}
                                                    & batch size $N_{t}$   & 10       \\ \cline{2-3}
									                & group size $n$       & 5        \\ \cline{2-3}
& gradient clip & 5.0     \\   \cline{2-3}
\hline
\end{tabular}
\end{center}
\caption{\label{hyper}Hyper-parameters of the models built in our experiments.}
\end{table}

\section{Experiment}
\subsection{Dataset and Evaluation Metrics}
The New York Times (NYT) dataset was adopted in our experiments.
This dataset was first released by \cite{riedel2010modeling} and has been widely used by previous research on distant supervision relation extraction \cite{liu2017soft,jat2018improving,qin2018dsgan,qin2018robust}.
This dataset was generated by aligning Freebase with the New York Times (NYT) corpus automatically.
There were 52 actual relations and a special relation ${NA}$ which indicated there was no relation between  two entities.

Following previous studies \citep{mintz2009distant,liu2017soft}, we evaluated our models on the held-out test set of the NYT dataset.
Precision-recall (PR) curves, area under curve (AUC) values and Precision@N (P@N) values \citep{lin2016neural} were adopted as evaluation metrics in our experiments.
All of the numerical results given by our experiments were the mean values of 10 repetitive trainings, and the PR curves were randomly selected from the repetitions
because there was no significant visual difference among them.

\subsection{Training Details and Hyperparameters}
All of the hyperparameters used in our experiments are listed in Table \ref{hyper}.
Most of them followed the hyperparameter settings in \citep{lin2016neural}.
The 50-dimensional word embeddings released by \citep{lin2016neural}\footnote{\url{https://github.com/thunlp/NRE}.} were also adopted for initialization.
The vocabulary contained the words which appeared more than 100 times in the NYT corpus.

Two different batch sizes $N_{p}$ and $N_{t}$ were used for pre-training and training respectively.
In our experiments, a grid search is employed  using training set to determine the optimal values of $n$, $N_p$ and $N_t$ among $n \in \{3, 4, ..., 10\}$, $N_{p} \in \{10, 20, 50, 100, 200\}$ and $N_{t} \in \{5, 10, 20, 50\}$.
Note that increasing the bag group size $n$ may boost the effect of inter-bag attentions but lead to less training samples.
The effects of inter-bag attentions would be lost when $n$=1.
For optimization, we employed mini-batch SGD with the initial learning rate of 0.1.
The learning rate was decayed to one tenth every 100,000 steps.
The pre-trained model with only intra-bag attentions converged within 300,000 steps in our experiments.
Thus, the initial learning rate for training the model with inter-bag attentions was set as 0.001.

\begin{figure}[!t]
\centering
\includegraphics[width=3.1in]{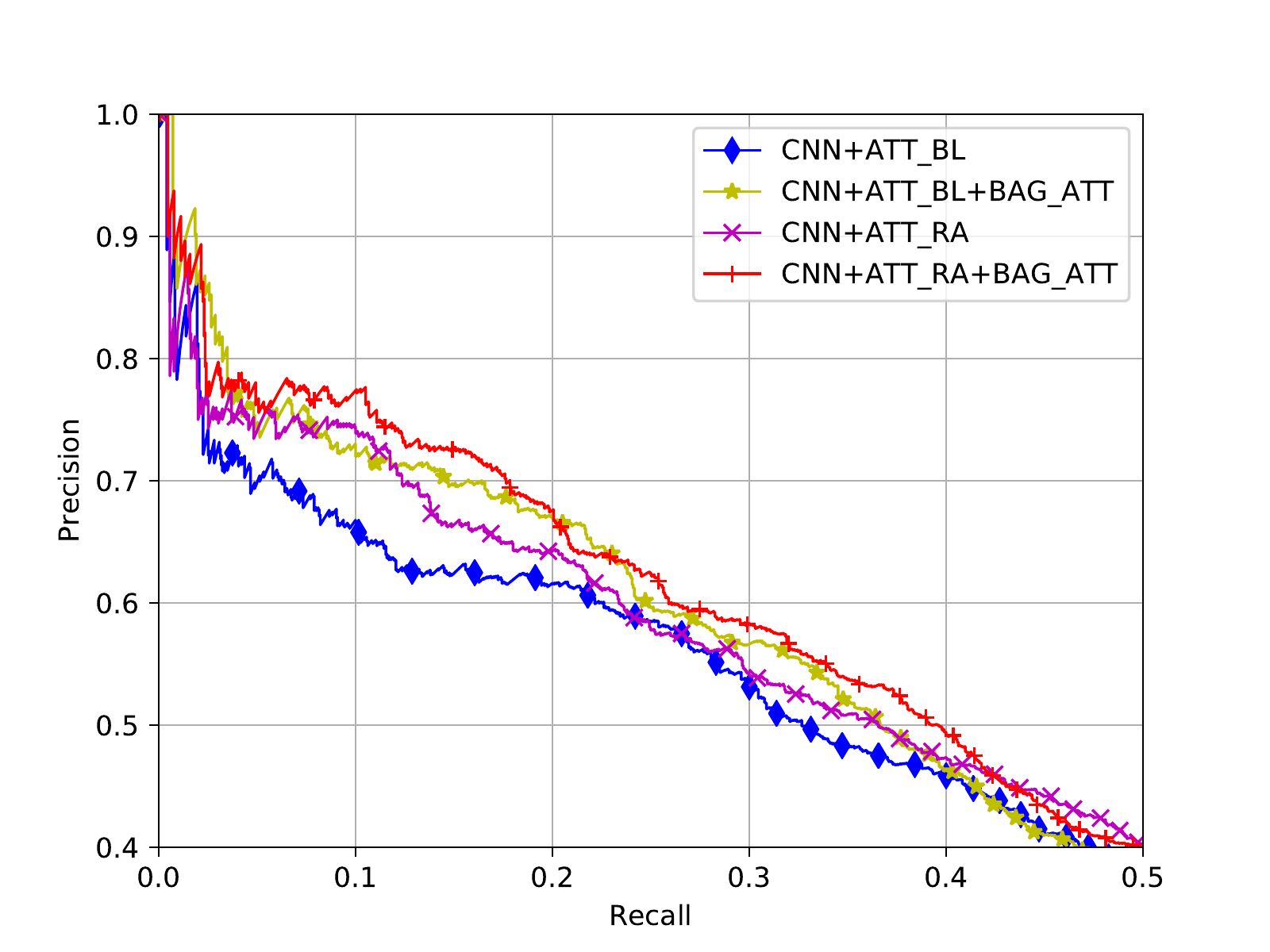}
\caption{PR curves of different models using CNN sentence encoders.}
\label{CNN}
\end{figure}

\subsection{Overall performance}
\begin{figure}[!t]
\centering
\includegraphics[width=3.1in]{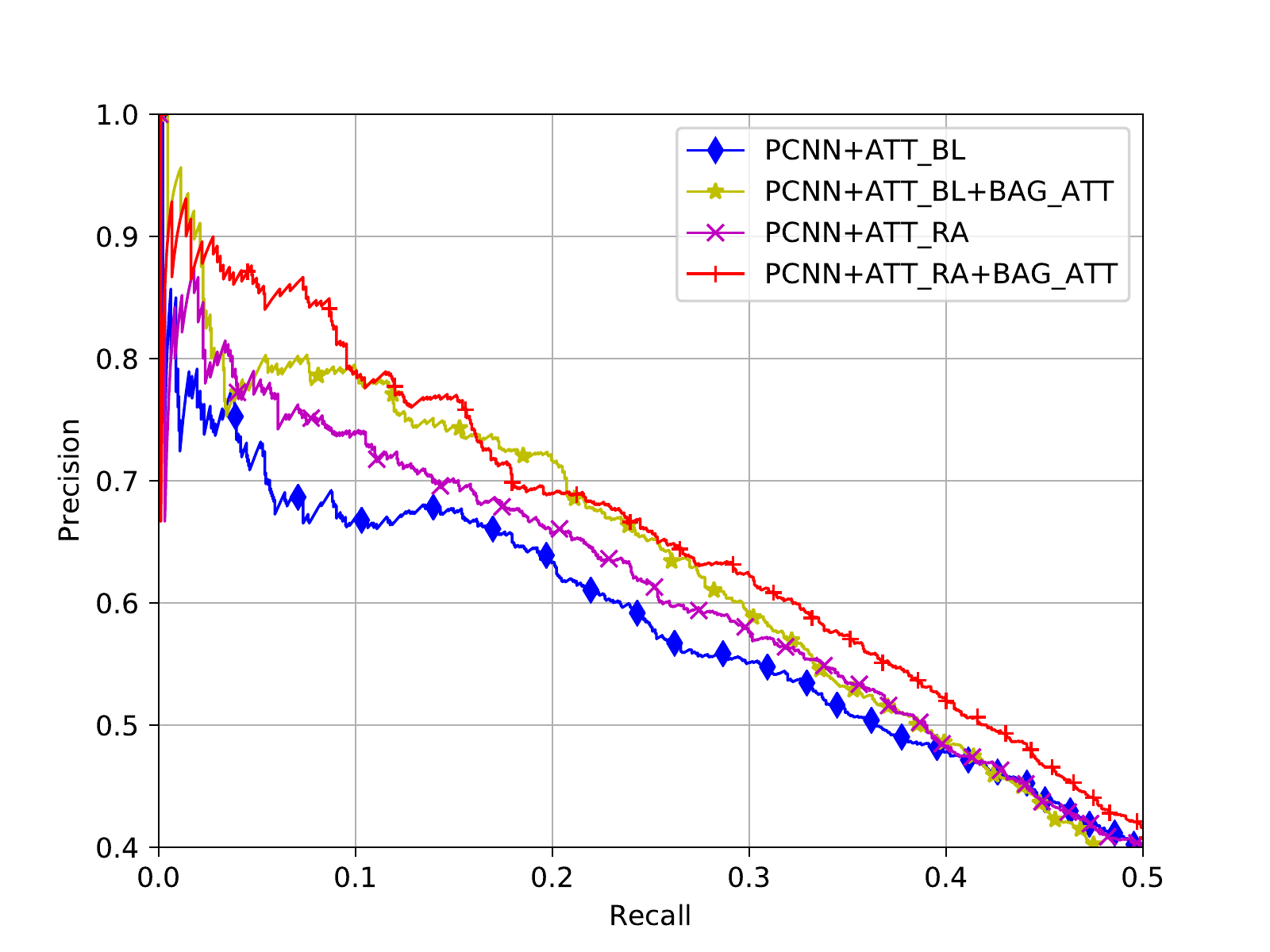}
\caption{PR curves of different models using PCNN sentence encoders.}
\label{PCNN}
\end{figure}

Eight models were implemented for comparison. The names of these models are listed in Table \ref{AUC},
where \emph{CNN} and \emph{PCNN} denote using CNNs or piecewise CNNs in sentence encoders respectively,
\emph{ATT\_BL} means the baseline intra-bag attention method proposed by \citep{lin2016neural},
\emph{ATT\_RA} means our proposed relation-aware intra-bag attention method,
and \emph{BAG\_ATT} means our proposed inter-bag attention method.
At the training stage of the  \emph{ATT\_BL} method, the relation query vector for attention weight calculation was fixed as the embedding vector associated with the distant supervision label for each bag.
At the test stage, all relation query vectors were applied to calculate the posterior probabilities of relations respectively and the relation with the highest probability was chosen as the classification result \citep{lin2016neural}.
The means and standard deviations of the AUC values given by the whole PR curves of these models are shown in Table \ref{AUC} for a quantitative comparison.
Following \citep{lin2016neural}, we also plotted the PR curves of these models in Fig. \ref{CNN} and \ref{PCNN} with recall smaller than $0.5$ for a visualized comparison.

\begin{table}[t!]
\small
\begin{center}
\begin{tabular}{|c|c|c|}
\hline
\bf No. & \bf Model    & \bf AUC \\ \hline
1 &CNN+ATT\_BL            & $0.376 \pm 0.003$ \\ \hline
2 &CNN+ATT\_BL+BAG\_ATT   & $0.388 \pm 0.002$\\ \hline
3 &CNN+ATT\_RA            & $0.398 \pm 0.004$ \\ \hline
4 &CNN+ATT\_RA+BAG\_ATT   & $0.407 \pm 0.004$\\ \hline
\hline
5 &PCNN+ATT\_BL           & $0.388 \pm 0.004$ \\ \hline
6 &PCNN+ATT\_BL+BAG\_ATT  & $0.403 \pm 0.002$ \\ \hline
7 &PCNN+ATT\_RA           & $0.403 \pm 0.003$ \\ \hline
8 &PCNN+ATT\_RA+BAG\_ATT  & \bf  0.422 $\pm$ 0.004 \\ \hline
\end{tabular}
\end{center}
\caption{ AUC values of different models. }
\label{AUC}
\end{table}

\begin{table*}
\begin{center}
\small
\begin{tabular}{|c|c|c|c|c|c|c|c|c|c|c|c|c|}
\hline
\bf \# of Test Sentences &\multicolumn{4}{c|}{\bf one} & \multicolumn{4}{c|}{\bf two} & \multicolumn{4}{c|}{\bf all} \\ \hline
P@N(\%) & 100 &200& 300 &mean& 100 &200& 300 &mean& 100 &200& 300 &mean \\ \hline
\cite{lin2016neural} &73.3&69.2&60.8&67.8&77.2&71.6&66.1&71.6&76.2&73.1&67.4&72.2 \\ \hline
\cite{liu2017soft}  &84.0&75.5&68.3&75.9&86.0&77.0&73.3&78.8&87.0&84.5&77.0&82.8  \\ \hline
\hline
CNN+ATT\_BL          &74.2 &68.9 &65.3 &69.5 &77.8 &71.5 &68.1 &72.5 &79.2 &74.9 &70.3 &74.8 \\ \hline
CNN+ATT\_RA          &76.8 &72.7 &67.9 &72.5 &79.6 &73.9 &70.7 &74.7 &81.4 &76.3 &72.5 &76.8 \\ \hline
CNN+ATT\_BL+BAG\_ATT &78.6 &74.2 &69.7 &74.2 &82.4 &76.2 &72.1 &76.9 &83.0 &78.0 &74.0 &78.3 \\ \hline
CNN+ATT\_RA+BAG\_ATT &79.8 &75.3 &71.0 &75.4 &83.2 &76.5 &72.1 &77.3 &87.2 &78.7 &74.9 &80.3 \\ \hline
\hline
PCNN+ATT\_BL          &78.6 &73.5 &68.1 &73.4 &77.8 &75.1 &70.3 &74.4 &80.8 &77.5 &72.3 &76.9 \\ \hline
PCNN+ATT\_RA          &79.4 &73.9 &69.6 &74.3 &82.2 &77.6 &72.4 &77.4 &84.2 &79.9 &73.0 &79.0 \\ \hline
PCNN+ATT\_BL+BAG\_ATT &85.2 &\bf 78.2 &71.3 &78.2 &84.8 &\bf 80.0 &74.3 &79.7 &88.8 &83.7 &77.4 &83.9 \\ \hline
PCNN+ATT\_RA+BAG\_ATT &\bf 86.8&77.6&\bf 73.9&\bf 79.4&\bf 91.2& 79.2&\bf 75.4&\bf 81.9&\bf 91.8&\bf 84.0&\bf 78.7&\bf 84.8 \\ \hline
\end{tabular}
\end{center}
\caption{P@N values of the entity pairs with different number of test sentences.}
\label{PN}
\end{table*}

From Table \ref{AUC}, Fig. \ref{CNN} and Fig. \ref{PCNN}, we have the following observations.
(1) Similar to the results of previous work \cite{zeng2015distant}, PCNNs worked better than CNNs as sentence encoders.
(2) When using either CNN or PCNN sentence encoders, \emph{ATT\_RA} outperformed \emph{ATT\_BL}.
It can be attributed to that the \emph{ATT\_BL} method only considered the target relation when deriving bag representations at training time,
while the \emph{ATT\_RA} method calculated intra-bag attention weights using all relation embeddings as queries, which improved the flexibility of bag representations.
(3) For both sentence encoders and both intra-bag attention methods, the models with \emph{BAG\_ATT} always achieved better performances than the ones without \emph{BAG\_ATT}.
This result verified the effectiveness of our proposed inter-bag attention method for distant supervision relation extraction.
(4) The best AUC performance was achieved by combining PCNN sentence encoders with the  intra-bag and inter-bag attentions proposed in this paper.

\begin{figure}[!t]
\centering
\includegraphics[width=3.1in]{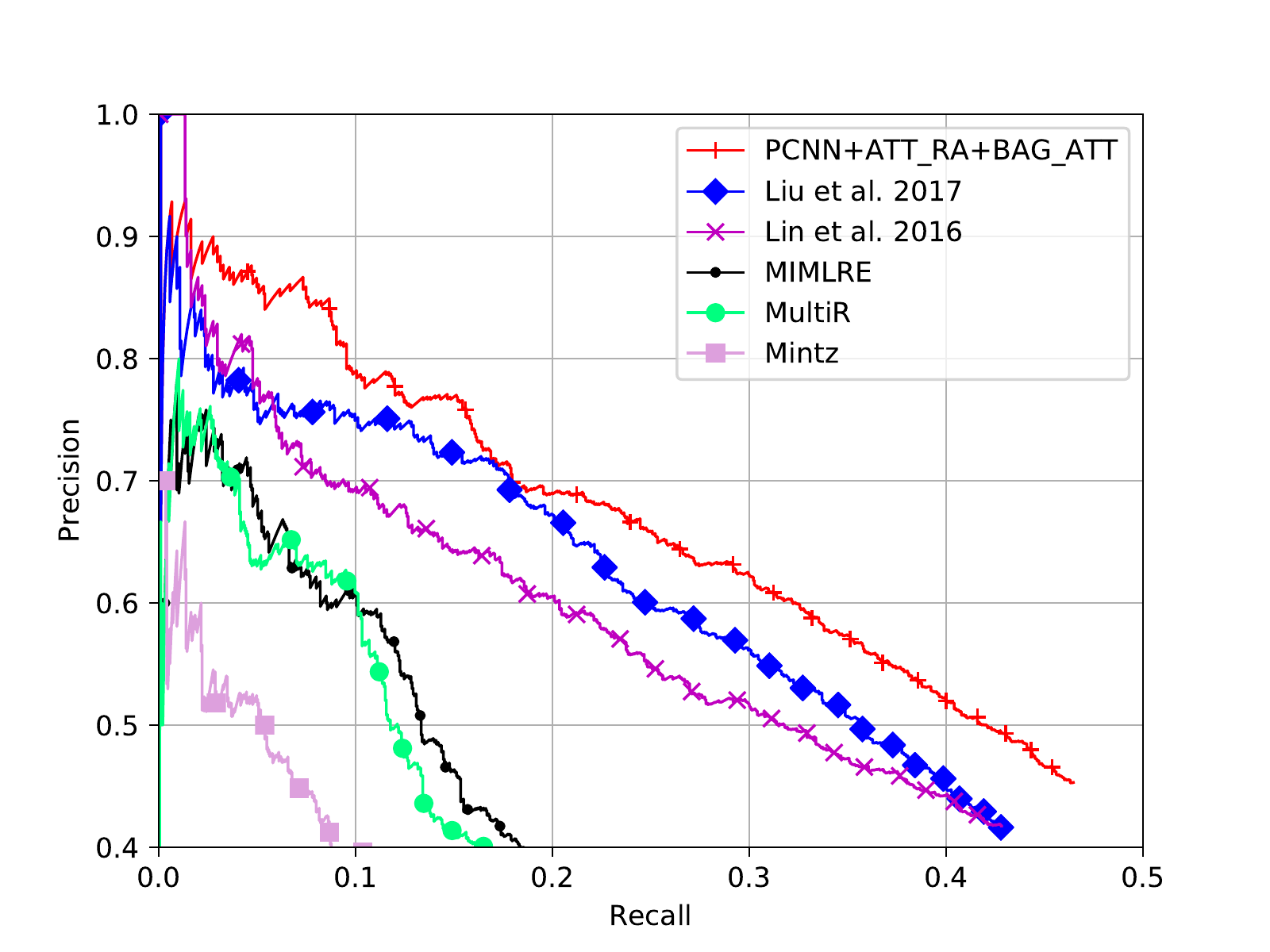}
\caption{PR curves of several models in previous work and our best model.}
\label{compare}
\end{figure}

\subsection{Comparison with previous work}
\subsubsection{PR curves}
The PR curves of several models in previous work and our best model \emph{PCNN+ATT\_RA+BAG\_ATT} are compared in Fig. \ref{compare},
where Mintz \cite{mintz2009distant}, MultiR \cite{hoffmann2011knowledge} and MIMLRE \cite{surdeanu2012multi} are conventional feature-based methods,
and \cite{lin2016neural} and \cite{liu2017soft} are PCNN-based ones\footnote{All of these curve data are from \url{https://github.com/tyliupku/soft-label-RE}.}.
For a fair comparison with \cite{lin2016neural} and \cite{liu2017soft}, we also plotted the curves with only the top 2000 points.
We can see that our model achieved better PR performance than all the other models.

\subsubsection{AUC values}
\emph{ATT\_BL+DSGAN} \cite{qin2018dsgan} and \emph{ATT\_BL+RL} \cite{qin2018robust} are two recent studies on distant supervision relation extraction with reinforcement learning for data filtering,
which reported the AUC values of PR curves composed by the top 2000 points.
Table \ref{comparewithrl} compares the AUC values reported in these two papers and the results of our proposed models.
We can see that introducing the proposed  \emph{ATT\_RA} and \emph{BAG\_ATT} methods to baseline models achieved larger improvement than using the methods proposed in \cite{qin2018dsgan,qin2018robust}.

\begin{table}[t!]
\small
\begin{center}
\begin{tabular}{|c|c|c|}
\hline
\bf model & \bf CNN & \bf PCNN \\ \hline
\bf ATT\_BL$^\dag$   & 0.219 & 0.253 \\ \hline
\bf ATT\_BL+RL       & 0.229 & 0.261 \\ \hline
\bf ATT\_BL+DSGAN    & 0.226 & 0.264 \\ \hline
\hline
\bf ATT\_BL$^\ddag$  & 0.242 & 0.271 \\ \hline
\bf ATT\_RA          & 0.254 & 0.297 \\ \hline
\bf ATT\_BL+BAG\_ATT & 0.253 & 0.285 \\ \hline
\bf ATT\_RA+BAG\_ATT & \bf 0.262 & \bf 0.311 \\ \hline
\end{tabular}
\end{center}
\caption{AUC values of previous work and our models, where \emph{ATT\_BL+DSGAN} and \emph{ATT\_BL+RL} are two models proposed in \cite{qin2018dsgan} and  \cite{qin2018robust} respectively,
$\dag$ indicates the baseline result reported in \cite{qin2018dsgan,qin2018robust} and $\ddag$ indicates the baseline result given by our implementation.}
\label{comparewithrl}
\end{table}

\begin{table*}[t!]
\small
\begin{center}
\begin{tabular}{|c|p{7.5cm}|c|c|c|}
\hline
\bf bag &\bf sentence  &\bf correct?& \bf intra-bag weights  & \bf inter-bag weights \\ \hline
 {\multirow{5}{*}{\bf B1}} & [\textbf{Panama City Beach}]$_{e2}$ , too , has a glut of condos , but the area was one of only two in [\textbf{Florida}]$_{e1}$ where sales rose in march , compared with a year earlier. & {\multirow{3}{*}{\bf Yes}}&{\multirow{3}{*}{0.71}} & {\multirow{5}{*}{0.48}}     \\ \cline{2-4}
& Like much of [\textbf{Florida}]$_{e1}$ , [\textbf{Panama City Beach}]$_{e2}$ has been hurt by the downturn in the real estate market.    &{\multirow{2}{*}{\bf Yes}} &{\multirow{2}{*}{0.29}}   &    \\ \hline
{\multirow{4}{*}{\bf B2}} & Among the major rivers that overflowed were the Housatonic , Still , Saugatuck , Norwalk , Quinnipiac , Farmington , [\textbf{Naugatuck}]$_{e1}$ , Mill , Rooster and [\textbf{Connecticut}]$_{e2}$ . &{\multirow{4}{*}{\bf No}}&{\multirow{4}{*}{1.00}}& {\multirow{4}{*}{0.13}} \\ \hline
{\multirow{6}{*}{\bf B3}}&..., the army chose a prominent location in [\textbf{Virginia}]$_{e1}$, at the foot of the Arlington memorial bridge , directly across the [\textbf{Potomac River}]$_{e2}$ from the Lincoln memorial . &{\multirow{3}{*}{\bf No}}& {\multirow{3}{*}{0.13}} & {\multirow{6}{*}{0.39}}     \\ \cline{2-4}
&... , none of those stars carried the giants the way barber did at Fedex field , across the [\textbf{Potomac River}]$_{e1}$ from [\textbf{Virginia}]$_{e2}$ , where he grew up as a redskins fan .   & {\multirow{3}{*}{\bf Yes}} & {\multirow{3}{*}{0.87}}   &    \\ \hline
\end{tabular}
\end{center}
\caption{A test set example of relation $\emph{/location/location/contains}$ from the NYT corpus. }
\label{bagattcase}
\end{table*}

\begin{table}[t!]
\small
\begin{center}
\begin{tabular}{|c|c|}
\hline
\bf sentence number & \bf mean$\pm$std \\ \hline
\bf 1 &$0.163\pm 0.029$  \\ \hline
\bf 2 &$0.187\pm 0.033$ \\ \hline
\bf 3 &$0.210\pm 0.034$ \\ \hline
\bf 4 &$0.212\pm 0.037$ \\ \hline
$\mathbf{\geq 5}$ &$0.256\pm 0.043$ \\ \hline
\end{tabular}
\end{center}
\caption{The distributions of inter-bag attention weights for the bags with different number of sentences. }
\label{weightdistribution}
\end{table}

\subsection{Effects of Intra-Bag Attentions}
Following \cite{lin2016neural}, we evaluated our models on the entity pairs with more than one training sentence.
One, two and all sentences for each test entity pair were randomly selected to construct three new test sets.
The P@100, P@200, P@300 values and their means given by our proposed models on these three test sets are reported in Table \ref{PN}
together with the best results of \cite{lin2016neural} and \cite{liu2017soft}.
Here, P@N means the precision of the relation classification results with the top $N$ highest probabilities in the test set.

We can see our proposed methods achieved higher P@N values than previous work.
Furthermore, no matter whether \emph{PCNN} or \emph{BAG\_ATT} were adopted, the \emph{ATT\_RA} method outperformed the  \emph{ATT\_BL} method on the test set with only one sentence for each entity pair.
Note that the decoding procedures of \emph{ATT\_BL} and \emph{ATT\_RA} were equivalent when there was only one sentence in a bag.
Therefore, the improvements from \emph{ATT\_BL} to \emph{ATT\_RA} can be attributed to that \emph{ATT\_RA} calculated intra-bag attention weights in a relation-aware way at the training stage.

\subsection{Distributions of Inter-Bag Attention Weights}
We divided the training set into 5 parts according to the number of sentences in each bag.
For each bag, the inter-bag attention weights given by the \emph{PCNN+ATT\_RA+BAG\_ATT} model were recorded.
Then, the mean and standard deviation of inter-bag attention weights for each part of the training set were calculated and are shown  in Table \ref{weightdistribution}.
From this table, we can see that the bag with smaller number of training sentences were usually assigned with lower inter-bag attention weights.
This result was consistent with the finding in \cite{qin2018robust} that the entity pairs with fewer training sentences were more likely to have incorrect relation labels.

\subsection{Case Study}
A test set example of relation $\emph{/location/location/contains}$ is shown in Table \ref{bagattcase}. The bag group contained 3 bags, which consisted of 2, 1, and 2 sentences respectively.
We calculated the intra-bag and inter-bag attentions for this bag group using our \emph{PCNN+ATT\_RA+BAG\_ATT} model and the weights of the target relation are also shown in Table \ref{bagattcase}.

In this example, the second bag was a noisy bag because the only sentence in this bag didn't express the relation $\emph{/location/location/contains}$ between the two entities \textbf{Naugatuck} and \textbf{Connecticut}.
In conventional methods, these three bags were treated equally for model training.
After introducing inter-bag attention mechanism, the weight of this noisy bag was reduced significantly as shown in the last column of Table \ref{bagattcase}.

\section{Conclusion}
In this paper, we have proposed a neural network with intra-bag and inter-bag attentions to cope with the noisy sentence and noisy bag problems in distant supervision relation extraction.
First, relation-aware bag representations are calculated by a weighted sum of sentence embeddings where the noisy sentences are expected to have smaller weights.
Further, an inter-bag attention module is designed to deal with the noisy bag problem by calculating the bag-level attention weights dynamically during model training.
Experimental results on  New York Times dataset show that our models achieved significant and consistent improvements compared with the models using only conventional intra-bag attentions.
To deal with the multi-label problem of relation extraction and to integrate external knowledge into our model will be  the tasks of our future work.

\section*{Acknowledgments}
We thank the anonymous reviewers for their valuable comments.

\bibliography{naaclhlt2019}
\bibliographystyle{acl_natbib}

\end{document}